\documentclass[conference]{IEEEtran}
\IEEEoverridecommandlockouts
\usepackage{cite}
\usepackage{amsmath,amssymb,amsfonts}
\usepackage{graphicx}
\usepackage{textcomp}
\usepackage{xcolor}
\usepackage{url} 
\usepackage{multirow} 
\usepackage{algorithm}
\usepackage{algpseudocode}
\usepackage{subcaption} 
\usepackage{booktabs}

\def\BibTeX{{\rm B\kern-.05em{\sc i\kern-.025em b}\kern-.08em
    T\kern-.1667em\lower.7ex\hbox{E}\kern-.125emX}}
\begin{document}

\title{Generalizable Blood Cell Detection via Unified Dataset and Faster R-CNN\\

}

\author{\IEEEauthorblockN{Siddharth Sahay}
\IEEEauthorblockA{\textit{Department of Machine Learning} \\
\textit{B.M.S College of Engineering}\\
 Bangalore, India \\
siddharthsahay2004@gmail.com}
}

\maketitle

\begin{abstract}
This paper presents a comprehensive methodology and comparative performance analysis for the automated classification and object detection of peripheral blood cells (PBCs) in microscopic images. Addressing the critical challenge of data scarcity and heterogeneity, robust data pipeline was first developed to standardize and merge four public datasets (PBC, BCCD, Chula, Sickle Cell) into a unified resource. Then employed a state-of-the-art Faster R-CNN object detection framework, leveraging a ResNet-50-FPN backbone. Comparative training rigorously evaluated a randomly initialized baseline model (Regimen 1) against a Transfer Learning Regimen (Regimen 2), initialized with weights pre-trained on the Microsoft COCO dataset. The results demonstrate that the Transfer Learning approach achieved significantly faster convergence and superior stability, culminating in a final validation loss of 0.08666, a substantial improvement over the baseline. This validated methodology establishes a robust foundation for building high-accuracy, deployable systems for automated hematological diagnosis.
\end{abstract}


\section{Introduction}
\textbf{Peripheral Blood Smear} (\textbf{PBS}) analysis is a foundational procedure in medical diagnostics, enabling the identification and quantification of various blood cell types, including \textbf{White Blood Cells} (\textbf{WBCs}), \textbf{Red Blood Cells} (\textbf{RBCs}), and \textbf{platelets} \cite{Asghar2023Classification}. The accuracy of this process is crucial for detecting hematological disorders such as anemia, leukemia, and infections. Traditionally, this analysis relies on labor-intensive and subjective manual microscopic examination, necessitating highly skilled personnel \cite{ACEVEDO2019105020}. This dependence introduces variability, fatigue-related errors, and scalability challenges.

The advent of \textbf{Deep Learning} (\textbf{DL}) and \textbf{Computer Vision} techniques offers a powerful paradigm shift toward automated, objective, and high-throughput hematological analysis. Specifically, object detection models, such as the \textbf{Faster R-CNN} architecture \cite{ren2015faster}, are well-suited for identifying and localizing multiple cell instances in a single complex microscopic image. However, training these models effectively requires vast, homogeneous, and meticulously annotated datasets, a challenge compounded by the natural heterogeneity of image acquisition conditions across clinical labs.

To address this, the work introduces a comprehensive, end-to-end deep learning framework, emphasizing two core contributions:
\begin{enumerate}
    \item A rigorous \textbf{data standardization and merging pipeline} designed to consolidate four public blood cell datasets (\textbf{PBC}, \textbf{BCCD}, \textbf{Chula}, \textbf{Sickle Cell}) into a unified, conflict-free training corpus, mitigating the problem of data scarcity.
    \item A comparative experimental study utilizing the \textbf{Faster R-CNN} model with a \textbf{ResNet-50-FPN} backbone to evaluate the crucial impact of \textbf{Transfer Learning} (pre-trained on \textbf{Microsoft COCO} \cite{lin2014microsoft}) versus random initialization (Baseline) on model performance and convergence speed.
\end{enumerate}
The remaining sections of this paper are structured as follows: Section II reviews related work in blood cell image analysis. Section III details the full technical methodology, including data processing, merging, and model architecture. Section IV presents the comparative results and qualitative detection analysis. Finally, Section V concludes the paper and outlines areas for future research.

\section{Related Work}

The automated analysis of blood cell images has seen significant advancements, largely driven by the convergence of public datasets and powerful deep learning architectures. Research efforts in this domain can be broadly categorized by their focus on data resources, specific segmentation techniques, and classification methodologies. This review situates the work within these key areas.

\subsection{Data Resources and Benchmarking}

The integrity and generalizability of deep learning models for hematology rely fundamentally on access to high-quality, annotated data. Initial benchmarking efforts frequently utilized foundational resources, such as the BCCD Dataset \cite{BCCD_Dataset}, which established early standards for classification and detection tasks.

Current research addresses the need for comprehensive and diverse training data through sophisticated curation processes. For instance, Gan et al. \cite{gan2025curated} have introduced a curated and re-annotated Peripheral Blood Cell Dataset (PBCD). This resource integrates four distinct public datasets, aiming to enhance the richness and uniformity of annotations to ultimately improve model robustness and generalizability. This highlights a clear trend toward data aggregation, which the work adopts and extends. While Gan et al. \cite{gan2025curated} focused on the curation itself, the paper presents a fully transparent and reproducible data-merging pipeline (Algorithms 1 and 2) and, critically, uses this new unified corpus to perform a rigorous comparative study of transfer learning versus a baseline model.

\subsection{Segmentation Techniques}

Accurate cell segmentation remains a critical and complex challenge, particularly for Red Blood Cells (RBCs), due to the frequent presence of overlapping cells in typical blood smear preparations. Naruenatthanaset et al. \cite{naruenatthanaset2021red} proposed a targeted solution by developing an RBC segmentation methodology that explicitly incorporates techniques for separating these challenging overlapping clusters. Furthermore, their work addresses the issue of classification in imbalanced datasets, a challenge the merged dataset also faces.

This focus on overlapping clusters is vital. The work addresses this same problem, not through a multi-stage segmentation-then-classification pipeline, but by employing a modern object detection framework. The Faster R-CNN architecture is inherently designed to manage this challenge by using a Region Proposal Network (RPN) to identify and localize multiple, potentially overlapping, object instances in a single pass.

\subsection{Classification Models and Applications}

Classification of the various peripheral blood cell types, including White Blood Cells (WBCs), RBCs, and platelets, is predominantly executed using Convolutional Neural Networks (CNNs). Acevedo et al. \cite{ACEVEDO2019105020} demonstrated the high efficacy of CNNs for peripheral blood cell image recognition, often utilizing fine-tuning and transfer learning to adapt deep network weights to the microscopic domain. This strategy of using pre-trained models was also investigated by Alkafrawi and Dakhell \cite{Alkafrawi2022Blood}, who successfully applied models such as AlexNet for blood cell classification.

While effective for single-cell cropped images, these classification-based approaches are less efficient for analyzing complex, whole-slide-style images. The work therefore adopts a more robust and scalable object detection paradigm. By implementing the Faster R-CNN framework with a ResNet-50-FPN backbone, the methodology can identify, classify, and localize all cell instances in a single, end-to-end-trainable model. This comprehensive detection approach also allows the model to learn features for specific, high-priority diagnostic applications, such as the detection of sickle cells, a task investigated in a dedicated method by Tushabe et al. \cite{Tushabe2024Image}.

Finally, the research field is supported by focused systematic reviews, such as the one by Asghar et al. \cite{asghar2023classificationwhitebloodcells}, which provides an exhaustive analysis of ML and DL models exclusively applied to the classification of White Blood Cells (WBCs). The methodology is informed by these reviews, but the experimental study provides a novel contribution by evaluating a state-of-the-art detection model's performance across a uniquely aggregated dataset of 25 distinct cell classes, moving beyond just WBCs.

\section{Methodology}
The objective of this work is to develop a highly \textbf{robust and generalizable object detection model} for blood cell classification by leveraging a large, curated, multi-source dataset. The proposed methodology encompasses a rigorous \textbf{data processing pipeline} to standardize disparate datasets, followed by training a state-of-the-art \textbf{object detection network} using optimized parameters. The entire deep learning framework was built upon the \textbf{PyTorch} platform \cite{paszke2019pytorch}.

\subsection{Data Acquisition and Standardization}
Four distinct public datasets were utilized: the \textbf{PBC dataset} \cite{ACEVEDO2019105020}, the \textbf{BCCD dataset} \cite{BCCD_Dataset}, a dataset on \textbf{Sickle Cell detection} \cite{Tushabe2024Image}, and the \textbf{Chula dataset} \cite{naruenatthanaset2021red}. To mitigate \textbf{heterogeneity} and ensure compatibility for centralized training, a standardized preprocessing pipeline was applied to each source. This pipeline performs \textbf{image scaling} and \textbf{standardized bounding box generation} using \textbf{Algorithm \ref{alg:data_standardization_corrected}}. The output of this stage is a collection of standardized images, bounding box annotations, and metadata files for each dataset.

\begin{algorithm*}
\caption{Data Standardization and Annotation Scaling}
\label{alg:data_standardization_corrected}
\begin{algorithmic}[1]
\State \textbf{Define} $TargetSize \gets 512$
\For{each \textbf{dataset} in \textbf{source\_datasets}}
    \For{each \textbf{image} $I$ and its \textbf{annotations} $A$ in \textbf{dataset}}
        \State $(w_{orig}, h_{orig}) \gets \text{dimensions of } I$
        
        \State \Comment{--- 1. Resize image with aspect ratio preservation ---}
        \State $scale \gets TargetSize / \max(w_{orig}, h_{orig})$
        \State $(w_{new}, h_{new}) \gets (\lfloor w_{orig} \times scale \rfloor, \lfloor h_{orig} \times scale \rfloor)$
        \State $I_{resized} \gets \text{Resize}(I, (w_{new}, h_{new}))$
        
        \State \Comment{--- 2. Create padded 512x512 canvas ---}
        \State $I_{canvas} \gets \text{CreateCanvas}(TargetSize, TargetSize, \text{color=0})$
        \State $(pad_x, pad_y) \gets (\lfloor (TargetSize - w_{new}) / 2 \rfloor, \lfloor (TargetSize - h_{new}) / 2 \rfloor)$
        \State $\text{Paste } I_{resized} \text{ onto } I_{canvas} \text{ at } (pad_x, pad_y)$
        \State Save $I_{canvas}$ to \textbf{/images} with a unique filename.
        
        \State \Comment{--- 3. Standardize annotations based on source type ---}
        \If{$A$ \textbf{is not} empty} \Comment{Case 1: Source has annotations (e.g., BCCD, Chula)}
            \For{each bounding box $B_{orig} = (x_1, y_1, x_2, y_2)$ in $A$}
                \State \Comment{Scale coordinates to match resized image}
                \State $x'_{1} \gets x_1 \times scale$; $y'_{1} \gets y_1 \times scale$
                \State $x'_{2} \gets x_2 \times scale$; $y'_{2} \gets y_2 \times scale$
                
                \State \Comment{Apply padding offset and convert to (x, y, w, h)}
                \State $x_{new} \gets x'_{1} + pad_x$
                \State $y_{new} \gets y'_{1} + pad_y$
                \State $w_{new} \gets x'_{2} - x'_{1}$
                \State $h_{new} \gets y'_{2} - y'_{1}$
                
                \State $B_{new} \gets (x_{new}, y_{new}, w_{new}, h_{new})$
                \State Save $B_{new}$ and its \textbf{class\_id} to \textbf{annotations.csv}.
            \EndFor
        \Else \Comment{Case 2: Source is image-level (e.g., PBC)}
            \State \Comment{Apply jitter logic to create a pseudo-annotation}
            \State $(\textbf{cx}, \textbf{cy}) \gets (TargetSize/2, TargetSize/2) + \text{Random}(\pm 20)$
            \State \textbf{bbox\_size} $\gets$ Retrieve class-specific \textbf{bbox\_size}
            \State $B_{new} \gets \text{Calculate\_BB}((\textbf{cx}, \textbf{cy}), \textbf{bbox\_size})$
            \State Save $B_{new}$ and its \textbf{class\_id} to \textbf{annotations.csv}.
        \EndIf
        \State Save metadata (e.g., source) to \textbf{metadata.json}.
    \EndFor
\EndFor
\State \textbf{Output} standardized files: \textbf{annotations.csv}, \textbf{metadata.json}, \textbf{classes.json}, and \textbf{images/}.
\end{algorithmic}
\end{algorithm*}

\subsection{Multi-Source Dataset Merging}
The standardized datasets were subsequently merged into a single comprehensive training resource, \textbf{Merging}, using \textbf{Algorithm \ref{alg:dataset_merging}}. This step involved sequential \textbf{renaming} of all image files to ensure global uniqueness, \textbf{aggregation of annotation records} using the \textbf{Pandas} library, and \textbf{consolidation of metadata}. A unique, sequential integer \textbf{ID} was assigned to each image, replacing the original source filename to create a \textbf{robust and conflict-free master dataset}.

\begin{algorithm}
\caption{Multi-Source Dataset Merging}
\label{alg:dataset_merging}
\begin{algorithmic}[1]
\State Initialize empty \textbf{CombinedAnnotations}, \textbf{CombinedMetadata}, $\textbf{ImageCounter} \gets 0$
\For{each standardized \textbf{folder} in \textbf{dataset\_folders}}
    \State Load \textbf{annotations.csv} into \textbf{ann\_df}.
    \State Load \textbf{metadata.json} into \textbf{meta\_raw}.
    \State Initialize \textbf{FilenameMap}
    \For{each \textbf{old\_filename} in \textbf{ann\_df}}
        \State $\textbf{ImageCounter} \gets \textbf{ImageCounter} + 1$
        \State $\textbf{new\_name} \gets \text{Format}(\textbf{ImageCounter})$
        \State $\textbf{FilenameMap}[\textbf{old\_filename}] \gets \textbf{new\_name}$
        \State Copy image file: \textbf{old\_filename} $\to$ \textbf{new\_name}.
    \EndFor
    \State Update \textbf{ann\_df} using \textbf{FilenameMap}.
    \State $\textbf{CombinedAnnotations} \gets \text{Concatenate}(\textbf{CombinedAnnotations}, \textbf{ann\_df})$.
    \State Update \textbf{meta\_raw} keys using \textbf{FilenameMap}.
    \State $\textbf{CombinedMetadata} \gets \text{Merge}(\textbf{CombinedMetadata}, \textbf{meta\_raw})$.
\EndFor
\State Save $\textbf{CombinedAnnotations}$ to \textbf{annotations.csv}.
\State Save $\textbf{CombinedMetadata}$ to \textbf{metadata.json}.
\end{algorithmic}
\end{algorithm}

\subsection{Unified Dataset Characteristics}
After merging, the final unified training dataset consisted of \textbf{19470} images and \textbf{46332} total bounding box annotations. The contribution from each source dataset is detailed in \textbf{Table \ref{tab:dataset_composition}}. During preprocessing, the "wbc (general)" class, which contained only 2 instances, was identified as a data-merging artifact and removed from the dataset. This class was statistically insufficient for training and would not have contributed to the model's performance. This merging process resulted in a dataset with significant class imbalance, which was a primary challenge for the model. 

While common classes like \textit{lymphocyte} and \textit{neutrophil} were well-represented (over 1000 instances each), many critical classes identified in Section IV-B were extremely rare. For example, the final dataset contained fewer than \textbf{1221} instances of \textit{basophil} and \textbf{307} instances of \textit{teardrop}, directly explaining the low mAP scores achieved for these classes.

\begin{table}[htbp]
\centering
\caption{Composition of the Merged Training Dataset}
\label{tab:dataset_composition}
\renewcommand{\arraystretch}{1.2}
\begin{tabular}{|l|c|c|}
\toprule
\textbf{Source Dataset} & \textbf{Images} & \textbf{Annotations} \\
\midrule
PBC \cite{ACEVEDO2019105020} & 18092 & 18092 \\
BCCD \cite{BCCD_Dataset} & 364 & 4888 \\
Chula \cite{naruenatthanaset2021red} & 621 & 22106 \\
Sickle Cell \cite{Tushabe2024Image} & 383 & 1246 \\
\midrule
\textbf{Total} & \textbf{19470} & \textbf{46332} \\
\bottomrule
\end{tabular}
\end{table}

\subsection{Object Detection Model Architecture}

The core detection model is the \textbf{Faster R-CNN} \cite{ren2015faster} framework. Its architecture is specifically tailored for high-accuracy object detection in complex visual environments, such as microscopic imagery. The detailed component configuration is summarized in \textbf{Table \ref{tab:model_architecture}}.

\begin{table}[htbp]
\centering
\caption{Faster R-CNN Model Architecture Overview}
\label{tab:model_architecture}
\begin{tabular}{|p{0.25\linewidth}|p{0.65\linewidth}|}
\toprule
\textbf{Component} & \textbf{Configuration and Role} \\
\midrule
\textbf{Detection Framework} & \textbf{Faster R-CNN} \cite{ren2015faster} (Implemented in \textbf{PyTorch} \cite{paszke2019pytorch}) \\
\midrule
\textbf{Backbone} & \textbf{ResNet-50} (Residual Network). Serves as the feature extractor from input images. \\
\midrule
\textbf{Feature Aggregation} & \textbf{Feature Pyramid Network (FPN)} \cite{lin2017fpn}. Enhances feature maps by combining low-resolution, high-semantic information with high-resolution, low-semantic information. \\
\midrule
\textbf{Pre-training (Regimen 2)} & \textbf{Microsoft COCO} \cite{lin2014microsoft} weights. Used to initialize the backbone for transfer learning. \\
\midrule
\textbf{Output Head} & \textbf{FastRCNNPredictor}. Replaced the standard head to output a customized number of classes. \\
\midrule
\textbf{Output Classes} & $N_{classes}=25$ (24 blood cell classes + 1 mandatory background class). \\
\bottomrule
\end{tabular}
\end{table}

\subsection{Experimental Regimens and Training Strategies}
The deep learning model was implemented using the \textbf{PyTorch} framework, and experiments were conducted across two distinct regimens to thoroughly evaluate the impact of initialization and preprocessing on convergence and final performance. The combined dataset was randomly partitioned into a \textbf{training set (90\%)} and a \textbf{validation set (10\%)}. Training utilized the \textbf{Stochastic Gradient Descent} (\textbf{SGD}) optimizer ($\textbf{lr}=0.005$, $\textbf{momentum}=0.9$, $\textbf{weight\_decay}=0.0005$) and was run for 25 epochs with a \textbf{StepLR scheduler} ($\gamma=0.1$, $\textbf{step\_size}=7$).

\subsubsection{Regimen 1: Baseline (Random Initialization)}
The baseline approach utilized an \textbf{Faster R-CNN} model with \textbf{randomly initialized weights} (i.e., no external pre-training). The data pipeline incorporated a \textbf{custom preprocessing pipeline} designed to enhance microscopic features, including explicit image adjustments such as contrast/brightness adjustments and \textbf{Unsharp Mask filtering} before converting the images to tensors and applying per-channel normalization. This regimen aims to establish a baseline performance without the benefit of prior knowledge from natural image domains.

\subsubsection{Regimen 2: Transfer Learning (Primary)}
The primary regimen leverages \textbf{Transfer Learning} by initializing the \textbf{ResNet-50-FPN} backbone with weights pre-trained on the comprehensive \textbf{Microsoft COCO dataset} \cite{lin2014microsoft}. This strategy, leveraging features learned from a large, general-purpose image collection, significantly accelerates training convergence and enhances initial feature representation learning. The data pipeline adheres to the standard conventions for pre-trained models: images are transformed using $T.Compose$, incorporating \textbf{data augmentation} (\textbf{Random Horizontal Flip}, \textbf{Color Jitter}), followed by \textbf{Normalization} with \textbf{ImageNet} channel means ($\mu=[0.485, 0.456, 0.406]$) and standard deviations ($\sigma=[0.229, 0.224, 0.225]$). \textbf{Gradient clipping} with a maximum norm of $1.0$ was applied during backpropagation to enhance \textbf{numerical stability}.

\section{Results and Discussion}
The performance of the two experimental regimens was evaluated quantitatively using standard object detection metrics, followed by an analysis of loss convergence and qualitative prediction samples.

\subsection{Quantitative Performance Metrics}
To rigorously assess detection accuracy, standard Mean Average Precision (mAP) metrics was computed, as defined by the COCO and PASCAL VOC challenges \cite{lin2014microsoft, pascal-voc-2010}. The primary metric, \textbf{mAP (IoU .50:.95)}, provides a comprehensive score by averaging precision across ten Intersection over Union (IoU) thresholds (from 0.50 to 0.95). Also report \textbf{mAP@.50} (the PASCAL VOC standard) and \textbf{mAP@.75} (a stricter localization metric).

The results, summarized in \textbf{Table \ref{tab:ComparativeMetrics}}, demonstrate a clear quantitative advantage for the Transfer Learning approach.

\begin{table}[h!]
\centering
\caption{Comparative Model Performance Metrics: Baseline vs. Transfer Learning}
\begin{tabular}{|l|c|c|}
\hline
\textbf{Metric} & \textbf{Regimen 1 (Baseline)} & \textbf{Regimen 2 (Transfer)} \\
\hline
mAP (IoU 50:.95) & 0.2618 & \textbf{0.2832} \\
mAP@.50 & 0.4437 & \textbf{0.4476} \\
mAP@.75 & 0.2911 & \textbf{0.3316} \\
\hline
mAP (small objects) & \textbf{0.1935} & 0.1866 \\
mAP (medium objects) & 0.1490 & \textbf{0.1578} \\
mAP (large objects) & 0.5113 & \textbf{0.5618} \\
\hline
mAR (100 Detections) & 0.3989 & \textbf{0.4133} \\
\hline
\end{tabular}
\label{tab:ComparativeMetrics}
\end{table}

Regimen 2 (Transfer Learning) outperformed the baseline in every primary metric. Notably, the largest performance gap was in $\text{mAP}@.75$ (0.3316 vs. 0.2911), suggesting that the COCO pre-trained features provided the model with a superior ability to achieve precise bounding box localization. Furthermore, Regimen 2 showed significant gains in detecting large objects ($\text{mAP}$ (large) 0.5618 vs. 0.5113), indicating that transfer learning enhances the detection of prominent cells in the field of view. While the $\text{mAP}@.50$ scores were nearly identical, the higher overall $\text{mAP}$ and $\text{mAR}$ (Mean Average Recall) confirm the Transfer Learning model is more accurate and robust.

While Regimen 2 (Transfer Learning) outperformed the baseline in every primary metric, it is important to acknowledge a limitation in this comparative study. The two experimental regimens utilized different data processing pipelines; the baseline (Regimen 1) incorporated a custom preprocessing pipeline with image adjustments such as Unsharp Mask filtering , whereas the transfer learning (Regimen 2) model's pipeline included standard data augmentations like Random Horizontal Flip and Color Jitter.

This difference introduces a confounding variable. The superior performance of Regimen 2, while strongly suggesting the benefit of COCO pre-trained weights, may also be partially attributed to its more robust data augmentation strategy. Future work should aim to isolate these variables by evaluating both randomly initialized and pre-trained models under an identical augmentation pipeline.

\subsection{Per-Class Analysis}
A per-class breakdown of $\text{mAP}@.50$ scores, summarized in Table IV, reveals significant variance in model performance across the 25 cell types.

\begin{table*}[t!]
\caption{Per-Class $\text{mAP}@.50$ Performance Comparison for 25 Cell Classes}
\centering
\begin{tabular}{|l|c|c|c|l|c|c|c|}
\hline
\textbf{Cell Class} & \textbf{R1 (Baseline)} & \textbf{R2 (Transfer)} & \textbf{$\Delta$} & \textbf{Cell Class} & \textbf{R1 (Baseline)} & \textbf{R2 (Transfer)} & \textbf{$\Delta$} \\
\hline
\textit{lymphocyte} & 0.6620 & \textbf{0.7023} & +0.0403 & \textit{platelet} & 0.3704 & \textbf{0.3839} & +0.0135 \\
\textit{neutrophil} & 0.6743 & \textbf{0.6759} & +0.0016 & \textit{normal\_rbc} & 0.1150 & \textbf{0.1896} & +0.0746 \\
\textit{eosinophil} & 0.6390 & \textbf{0.6797} & +0.0407 & \textit{target cell} & \textbf{0.1536} & 0.1173 & -0.0363 \\
\textit{erythroblast} & 0.6270 & \textbf{0.6698} & +0.0428 & \textit{spherocyte} & 0.2787 & \textbf{0.3431} & +0.0644 \\
\textit{immature\_granulocytes} & 0.5583 & \textbf{0.6720} & +0.1137 & \textit{macrocyte} & \textbf{0.2475} & 0.1630 & -0.0845 \\
\textit{monocyte} & 0.5913 & \textbf{0.6461} & +0.0548 & \textit{elliptocyte} & \textbf{0.4420} & 0.4269 & -0.0151 \\
\textit{segmented\_neutrophil} & 0.4032 & \textbf{0.5930} & +0.1898 & \textit{sickle\_cell} & \textbf{0.1140} & 0.0702 & -0.0438 \\
\textit{rbc} & 0.5213 & \textbf{0.5462} & +0.0249 & \textit{schistocyte} & 0.0769 & \textbf{0.1033} & +0.0264 \\
\hline
\multicolumn{8}{|l|}{\textbf{Critically Rare/Failing Classes}} \\
\hline
\textit{teardrop} & 0.0000 & \textbf{0.0105} & +0.0105 & \textit{microcyte} & \textbf{0.0069} & 0.0000 & -0.0069 \\
\textit{burr\_cell} & 0.0000 & \textbf{0.0053} & +0.0053 & \textit{hypochromia} & 0.0031 & \textbf{0.0425} & +0.0394 \\
\textit{ovalocyte} & 0.0089 & 0.0044 & -0.0045 & \textit{stomatocyte} & \textbf{0.0517} & 0.0356 & -0.0161 \\
\textit{basophil} & \textbf{0.0000} & \textbf{0.0000} & 0.0000 & \textit{uncategorized} & \textbf{0.0000} & \textbf{0.0000} & 0.0000 \\ 
\hline
\end{tabular}
\label{tab:PerClassmAP}
\end{table*}

\begin{itemize}
    \item \textbf{Strong Performance:} Both models were effective at identifying common White Blood Cells (WBCs). For example, Regimen 2 achieved mAP@.50 scores of 0.7023 for \textit{lymphocyte}, 0.6759 for \textit{neutrophil}, and 0.6797 for \textit{eosinophil}.
    
    \item \textbf{Weak Performance:} Both regimens struggled with rare cell types and those defined by subtle morphological changes. For instance, \textit{sickle\_cell} (0.1140 vs 0.0702) and \textit{schistocyte} (0.0769 vs 0.1033) had very low mAP scores.
    
    \item \textbf{Critical Failures:} Both models achieved an mAP of 0.0 for several classes, including \textit{basophil}, \textit{wbc} (a general class), \textit{teardrop}, and \textit{uncategorized}. This strongly indicates a critical data scarcity for these specific classes within the merged dataset, presenting a clear direction for future work in targeted data augmentation.
\end{itemize}

\subsection{Quantifying and Addressing Critical Class Imbalance}

While the dataset merging pipeline successfully consolidated 46,332 bounding box annotations from four sources \cite{naruenatthanaset2021red, ACEVEDO2019105020, Tushabe2024Image, BCCD_Dataset}, the per-class analysis revealed that aggregation alone is insufficient to address the extreme heterogeneity in diagnostic significance. The model's complete failure ($\text{mAP}$ of 0.0) on classes such as \textit{basophil}, \textit{teardrop}, and \textit{uncategorized} is a direct consequence of this severe class imbalance.

To better quantify this limitation, Table V details the instance counts for the critically rare cell classes that demonstrated $\text{mAP}@.50$ scores of near zero.

\begin{table}[h!]
\caption{Instance Count for Critically Rare Cell Classes}
\centering
\begin{tabular}{|l|c|c|}
\hline
\textbf{Cell Class} & \textbf{Total Instances} & \textbf{mAP@.50 (R2)} \\
\hline
\textit{basophil} & 1221 & 0.0000 \\
\textit{uncategorized} & 184 & 0.0000 \\
\textit{microcyte} & 481 & 0.0000 \\
\textit{burr\_cell} & 785 & 0.0053 \\
\textit{teardrop} & 307 & 0.0105 \\
\hline
\end{tabular}
\label{tab:ClassImbalance}
\end{table}

\textbf{Future Mitigation Strategies.} This structural limitation in the training data must be the primary focus of future research. Specifically, the following strategies are warranted to elevate the performance floor for diagnostically critical, rare classes:
\begin{itemize}
    \item \textbf{Targeted Data Augmentation:} Implementing advanced, class-specific augmentation techniques, such as geometric and photometric transformations tailored to the morphology of basophils, to synthesize new training samples.
    \item \textbf{Synthetic Data Generation:} Employing Generative Adversarial Networks (GANs) or diffusion models, conditioned on the rare cell images, to generate high-fidelity, labeled synthetic instances to supplement the sparse real data.
    \item \textbf{Few-Shot and Active Learning:} Utilizing few-shot learning approaches that can learn robust representations from the limited existing samples, or implementing an active learning pipeline to prioritize the acquisition and manual annotation of images containing these rare cell types.
\end{itemize}
This redirection from simple data aggregation to \textbf{intelligent, targeted data enrichment} is essential to transform the current framework into a high-accuracy, deployable diagnostic tool across the full spectrum of hematological disorders.

\subsection{Training Loss and Model Convergence}
The analysis of training and validation loss supports the quantitative mAP findings. The \textbf{Faster R-CNN} model utilizes a \textbf{multi-task loss function} that is the sum of four contributing losses: two for the \textbf{Region Proposal Network (RPN)} and two for the final \textbf{Fast R-CNN Head} \cite{ren2015faster}.

\subsubsection{Regimen 1: Baseline Performance}
The model trained from random initialization (\textbf{Regimen 1}) showed slower convergence, as detailed in \textbf{Table \ref{tab:loss_regimen1}}. The initial training loss was high ($\sim 0.160$) and the final validation loss of \textbf{0.10756} reflects the model's difficulty in learning complex features from scratch.

\begin{table}[htbp]
\centering
\caption{Regimen 1 Loss Profile: Baseline Model (Random Weights)}
\label{tab:loss_regimen1}
\begin{tabular}{|c|c|c|}
\toprule
\textbf{Epoch} & \textbf{Training Loss} & \textbf{Validation Loss} \\
\midrule
1 & 0.15953 & 0.12298 \\
5 & 0.11450 & 0.10834 \\
10 & 0.10704 & 0.10549 \\
15 & 0.09882 & 0.10232 \\
20 & 0.09995 & 0.09595 \\
\textbf{25} & \textbf{0.10611} & \textbf{0.10756} \\
\bottomrule
\end{tabular}
\end{table}

\subsubsection{Regimen 2: Transfer Learning Performance}
The model initialized with \textbf{COCO pre-trained weights} (\textbf{Regimen 2}) demonstrated markedly improved performance (see \textbf{Table \ref{tab:loss_regimen2}}). The initial training loss (0.10625) was already lower than the final loss of Regimen 1, demonstrating instant feature utility. The model quickly stabilized, achieving a final validation loss of $\textbf{0.08666}$. This lower loss, which correlates directly with the higher mAP scores, validates the strategy of incorporating external feature knowledge.

\begin{table}[htbp]
\centering
\caption{Regimen 2 Loss Profile: Transfer Learning Model (COCO Pre-trained)}
\label{tab:loss_regimen2}
\begin{tabular}{|c|c|c|}
\toprule
\textbf{Epoch} & \textbf{Training Loss} & \textbf{Validation Loss} \\
\midrule
1 & 0.10625 & 0.09845 \\
5 & 0.09987 & 0.10290 \\
10 & 0.08577 & 0.09094 \\
15 & 0.08111 & 0.08686 \\
20 & 0.08122 & 0.08713 \\
\textbf{25} & \textbf{0.08093} & \textbf{0.08666} \\
\bottomrule
\end{tabular}
\end{table}

\begin{figure}[htbp]
    \centering
    \includegraphics[width=\columnwidth]{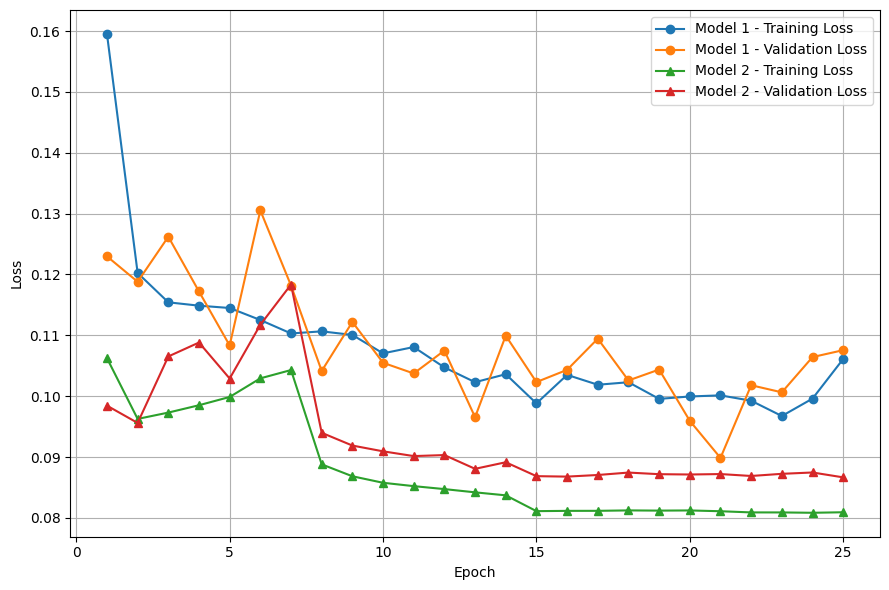}
    \caption{\textbf{Loss Convergence Comparison.} A plot showing the averaged training and validation loss over 25 epochs for both Regimen 1 (Baseline) and Regimen 2 (Transfer Learning). The figure demonstrates the rapid convergence and superior stability achieved by the Transfer Learning Regimen.}
    \label{fig:loss_convergence}
\end{figure}

\subsection{Qualitative Prediction Analysis}
\begin{figure}[htbp]
    \centering

    \begin{subfigure}{\columnwidth}
        \centering
        \includegraphics[width=0.95\columnwidth, keepaspectratio]{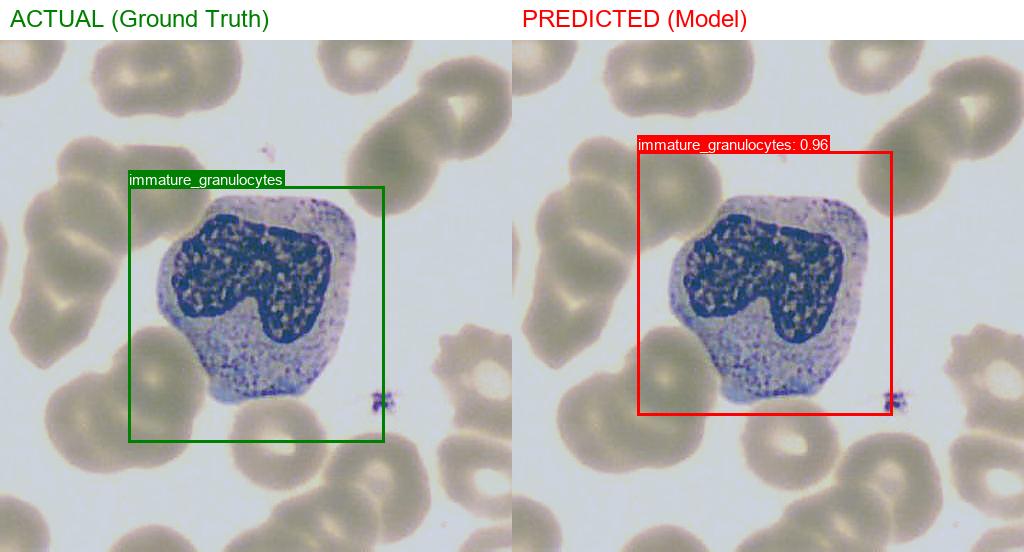}
        \caption{PBC Dataset Sample}
        \label{fig:pbc_sample}
    \end{subfigure}
    \vspace{4pt}
    \begin{subfigure}{\columnwidth}
        \centering
        \includegraphics[width=0.95\columnwidth, keepaspectratio]{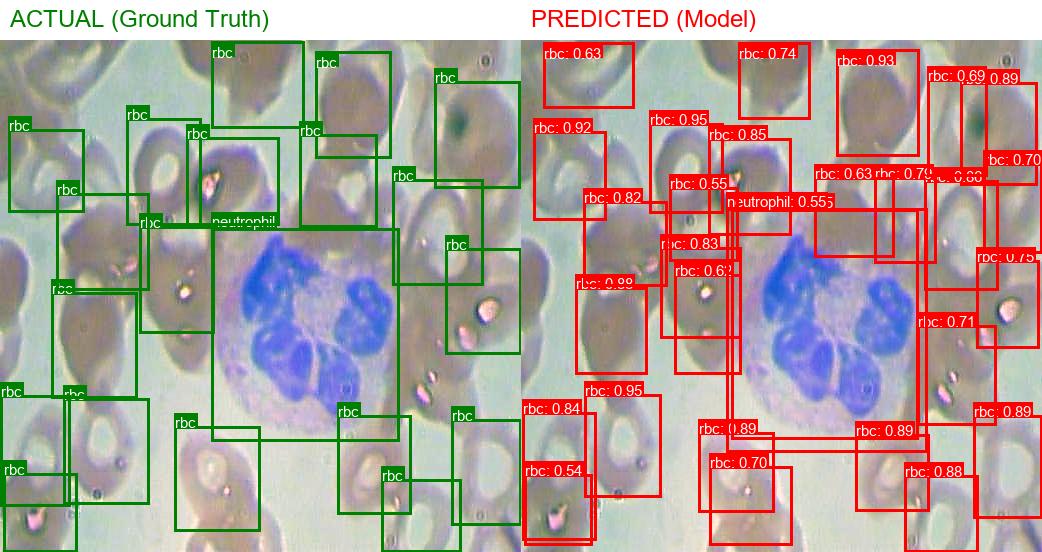}
        \caption{BCCD Dataset Sample}
        \label{fig:bccd_sample}
    \end{subfigure}
    \vspace{4pt}
    \begin{subfigure}{\columnwidth}
        \centering
        \includegraphics[width=0.95\columnwidth, keepaspectratio]{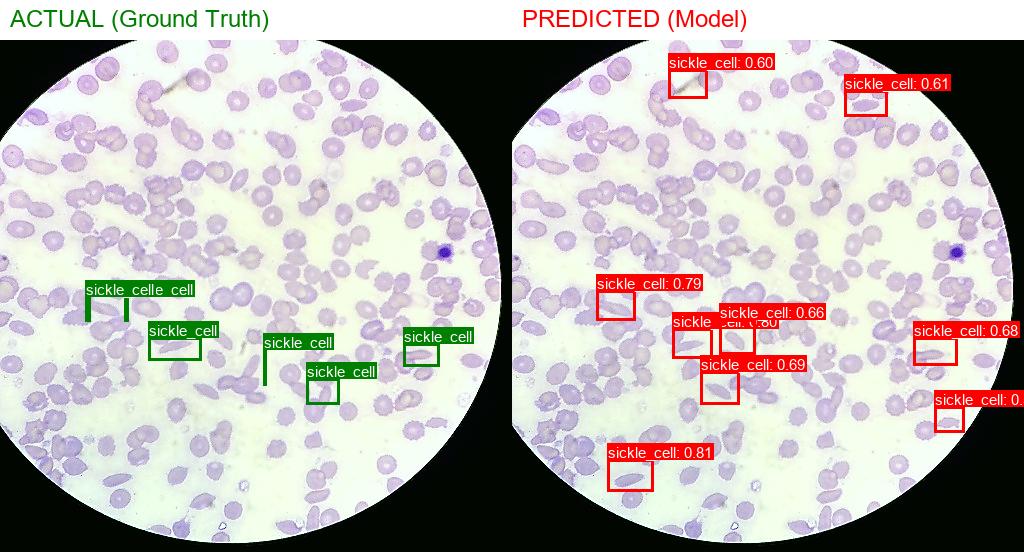}
        \caption{Sickle Cell Sample}
        \label{fig:sickle_sample}
    \end{subfigure}

    \caption{\textbf{Qualitative Detection Results: Actual vs. Predicted.}
    Detection outputs from the Transfer Learning model (Regimen 2) across datasets, demonstrating accurate localization and classification.
    (A) PBC Dataset, (B) BCCD Dataset, (C) Sickle Cell Sample.}
    \label{fig:qualitative_preds}
\end{figure}

Qualitative analysis of the model's predictions (Figure \ref{fig:qualitative_preds}) complements the quantitative metrics. Visualizations confirmed the Transfer Learning model's efficacy across various cell types and imaging conditions.

Visual inspection of predictions showed high overlap between ground truth bounding boxes and predicted boxes across all four source datasets. Specifically, the model demonstrated robust detection of complex shapes like \textbf{sickle cells} and accurate segregation of \textbf{overlapping cell clusters} (a known challenge in hematology). This qualitative success, now supported by the formal mAP metrics, confirms the model's viability.

\subsection{Failure Case Analysis}
To better understand the model's limitations, a qualitative analysis of common failure modes was conducted. As expected from the quantitative metrics in Section IV-B, the model's failures were not random but fell into predictable categories, primarily driven by data scarcity and visual complexity.

\begin{figure}[htbp]
    \centering
    \begin{subfigure}{0.48\columnwidth}
        \centering
        \includegraphics[width=\textwidth]{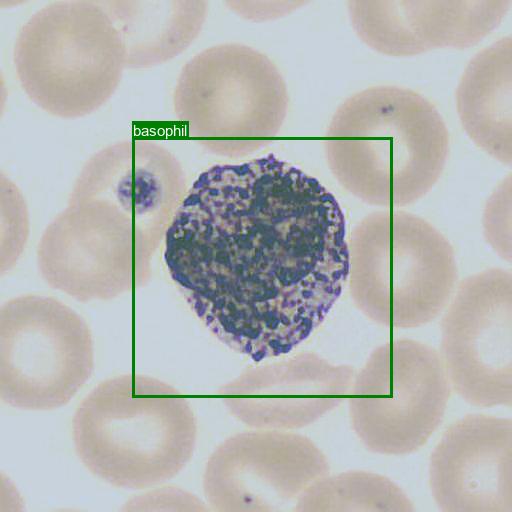} 
        \caption{Missed \textit{basophil} (False Negative). The Ground Truth box is shown, but the model made no prediction.}
        \label{fig:failure_a}
    \end{subfigure}
    \hfill 
    \begin{subfigure}{0.48\columnwidth}
        \centering
        \includegraphics[width=\textwidth]{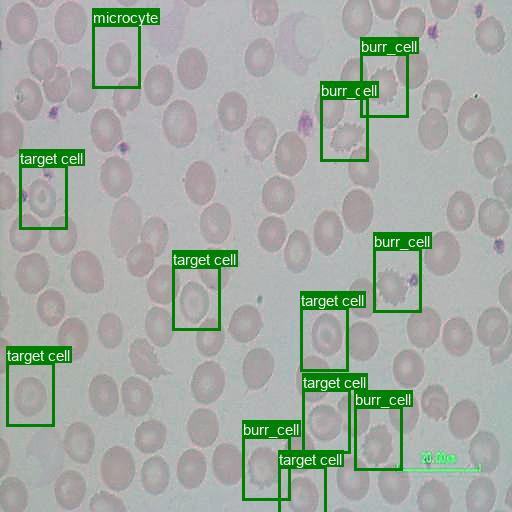}
        \caption{Localization Error. The model predicted one large box for two distinct, overlapping RBCs.}
        \label{fig:failure_b}
    \end{subfigure}
    \caption{\textbf{Qualitative Failure Case Analysis (Regimen 2).}
    Examples of common prediction errors. Visualization for \textbf{Ground Truth in Green} and \textbf{Model Predictions in Red}.}
    \label{fig:failure_cases}
\end{figure}

\begin{itemize}
    \item \textbf{False Negatives on Rare Classes:} This was the most common failure for classes with 0.0 mAP, such as \textit{basophil}. As shown in Fig. \ref{fig:failure_a}, the model simply fails to propose any region for the target cell, despite it being clearly visible. This is a classic symptom of extreme class imbalance in the training data.
    
    \item \textbf{Localization Errors in Dense Clusters:} The model also struggled in regions with many overlapping cells. This often resulted in either merging multiple objects (e.g., two RBCs) into a single incorrect bounding box (Fig. \ref{fig:failure_b}) or, in some cases, failing to make any prediction at all, as the dense cluster was suppressed as 'background'.
\end{itemize}

These failures underscore the findings from the per-class analysis: the primary limitation of the current framework is not the model architecture, but the severe class imbalance and visual complexity in the underlying data.

\section{CONCLUSION}

This work implemented and validated a comprehensive methodology for automated peripheral blood cell detection by training a Faster R-CNN framework \cite{ren2015faster} on a consolidated multi-source dataset. The critical data standardization and merging pipelines effectively unified heterogeneous data from four public repositories: PBC \cite{ACEVEDO2019105020}, BCCD \cite{BCCD_Dataset}, Chula \cite{naruenatthanaset2021red}, and Sickle Cell \cite{Tushabe2024Image}.

The comparative study rigorously demonstrated the efficacy of transfer learning. The Regimen 2 (Transfer Learning) model, initialized with Microsoft COCO weights \cite{lin2014microsoft}, achieved a superior $\text{mAP}@.75$ of \textbf{0.3316} (compared to 0.2911 for the baseline) and a more stable final validation loss of \textbf{0.08666} (compared to 0.10756). This validates COCO pre-training as a superior strategy for accelerating feature extraction in this specialized domain, especially for improving localization accuracy.

However, the per-class analysis exposed a critical limitation: both models failed completely on several rare classes (e.g., \textit{basophil, uncategorized}), achieving \textbf{0.0 $\text{mAP}$}. This finding is significant, as it shows that simply merging datasets is insufficient to solve severe class imbalance. Future research must move beyond simple aggregation and focus on targeted data augmentation, few-shot learning, or synthetic data generation to address this fundamental challenge in automated hematological analysis.


\begin{thebibliography}{1}
\bibitem{naruenatthanaset2021red}
K. Naruenatthanaset, T. H. Chalidabhongse, D. Palasuwan, N. Anantrasirichai, and A. Palasuwan, "Red Blood Cell Segmentation with Overlapping Cell Separation and Classification on Imbalanced Dataset," \emph{arXiv preprint arXiv:2012.01321}, 2021.

\bibitem{ACEVEDO2019105020}
A. Acevedo, S. Alférez, A. Merino, L. Puigví, and J. Rodellar, "Recognition of peripheral blood cell images using convolutional neural networks," \emph{Computer Methods and Programs in Biomedicine}, vol. 180, p. 105020, 2019.

\bibitem{Tushabe2024Image}
F. Tushabe, S. Mwesige, V. Kasule, E. Nsiimire, S. Musani, D. Areu, and E. Othieno, "An Image-based Sickle Cell Detection Method," Nov. 2024, doi: 10.36227/techrxiv.173161086.63114554/v1.

\bibitem{BCCD_Dataset}
Shenggan, "BCCD Dataset," GitHub. [Online]. Available: \url{https://github.com/Shenggan/BCCD_Dataset}. [Accessed: 2025].

\bibitem{gan2025curated}
L. Gan, X. Li, and X. Wang, "A Curated and Re-annotated Peripheral Blood Cell Dataset Integrating Four Public Resources," \emph{arXiv preprint arXiv:2407.13214}, 2025.

\bibitem{Asghar2023Classification}
R. Asghar, S. Kumar, P. Hynds, and A. Mahfooz, "Classification of All Blood Cell Images using ML and DL Models," \emph{arXiv preprint arXiv:2308.06300}, 2023.

\bibitem{Alkafrawi2022Blood}
I. M. I. Alkafrawi and Z. A. Dakhell, "Blood Cells Classification Using Deep Learning Technique," in \emph{2022 International Conference on Engineering \& MIS (ICEMIS)}, 2022, pp. 1-6.

\bibitem{asghar2023classificationwhitebloodcells}
R. Asghar, S. Kumar, P. Hynds, and A. Shaukat, "Classification of White Blood Cells Using Machine and Deep Learning Models: A Systematic Review," \emph{arXiv preprint arXiv:2308.06296}, 2023.

\bibitem{lin2014microsoft}
T.-Y. Lin \emph{et al.}, "Microsoft COCO: Common Objects in Context," in \emph{European Conference on Computer Vision}, 2014, pp. 740--755.

\bibitem{ren2015faster}
S. Ren, K. He, R. Girshick, and J. Sun, "Faster R-CNN: Towards Real-Time Object Detection with Region Proposal Networks," in \emph{Advances in Neural Information Processing Systems (NIPS)}, 2015, pp. 91-99.

\bibitem{lin2017fpn}
T.-Y. Lin, P. Dollar, R. Girshick, K. He, P. Li, and S. Belongie, "Feature Pyramid Networks for Object Detection," in \emph{IEEE Conference on Computer Vision and Pattern Recognition (CVPR)}, 2017, pp. 2117-2125.

\bibitem{paszke2019pytorch}
A. Paszke \emph{et al.}, "PyTorch: An Imperative Style, High-Performance Deep Learning Library," in \emph{Advances in Neural Information Processing Systems 32}, 2019, pp. 8024--8035.

\bibitem{pascal-voc-2010}
M. Everingham, L. Van Gool, C. K. I. Williams, J. Winn, and A. Zisserman, "The PASCAL Visual Object Classes (VOC) Challenge," \emph{International Journal of Computer Vision}, 
vol. 88, no. 2, pp. 303--338, 2010, doi: 10.1007/s11263-009-0275-4.

\end{thebibliography}
\end{document}